\documentclass{article}

\usepackage[final,nonatbib]{neurips_2020}

\usepackage[utf8]{inputenc} 
\usepackage[T1]{fontenc}    
\usepackage{booktabs}       
\usepackage{amsfonts}       
\usepackage{nicefrac}       
\usepackage{microtype}      
\usepackage{graphicx}
\usepackage{amsmath}
\usepackage{titlesec}
\usepackage[inline]{enumitem}

\setlist[enumerate,1]{label=\textbf{(\alph*)}}
\titleformat{\subsubsection}[runin]
{\normalfont}{\thesubsubsection}{1em}{}

\title{Bayesian Active Learning for Wearable Stress and Affect Detection}

\author{%
    Abhijith Ragav\thanks{Both authors contributed equally to this work.} \\
    Solarillion Foundation \\
    \texttt{abhijithragav@ieee.org} \\
    \And
    Gautham Krishna Gudur\footnotemark[1] \\
    Global AI Accelerator, Ericsson \\
    \texttt{gautham.krishna.gudur@ericsson.com} \\
}

\begin{document}

\maketitle

\sloppy

\begin{abstract}
In the recent past, psychological stress has been increasingly observed in humans, and early detection is crucial to prevent health risks. Stress detection using on-device deep learning algorithms has been on the rise owing to advancements in pervasive computing. However, an important challenge that needs to be addressed is handling unlabeled data in real-time via suitable ground truthing techniques (like Active Learning), which should help establish affective states (labels) while also selecting only the most informative data points to query from an oracle. In this paper, we propose a framework with capabilities to represent model uncertainties through approximations in Bayesian Neural Networks using Monte-Carlo (MC) Dropout. This is combined with suitable acquisition functions for active learning. Empirical results on a popular stress and affect detection dataset experimented on a Raspberry Pi 2 indicate that our proposed framework achieves a considerable efficiency boost during inference, with a substantially low number of acquired pool points during active learning across various acquisition functions. Variation Ratios achieves an accuracy of 90.38\% which is comparable to the maximum test accuracy achieved while training on about 40\% lesser data.
\end{abstract}

\maketitle

\section{Introduction}
Psychological stress is encountered by humans in response to physical, mental or emotional challenges presented to them by the environment. Physiological signals can be easily measured using wearable devices and are good indicators of stress, making them suitable for continuous stress monitoring. Deep learning architectures have recently performed well for such tasks, due to their automatic feature extraction capabilities, in contrast to conventional machine learning models which mandate domain knowledge to craft shallow heuristic features \cite{stress1}, \cite{stress_rnn}. There has been a special interest in bringing deep learning to wearable devices to provide real-time stress and affect monitoring, whilst preserving sensitive user data \cite{stress_abhijith}.

Typically, we assume that the physiological user data is inherently labeled, however it is almost practically never the case. One of the least explored areas involving deep learning for such tasks is \textit{Active Learning} -- a technique which gives a model, the ability to learn from real-world unlabeled data by querying an oracle (user). By using \textit{Bayesian Neural Networks (BNNs)}, which integrates \textit{Monte Carlo Dropout} with traditional neural networks \cite{dropout_yaringal}, it is possible to estimate predictive uncertainties. This is coupled with active learning \textit{acquisition functions} for querying the most uncertain data points from the oracle \cite{BayesianAL}. However, these works have not been considerably discussed in on-device mobile health scenarios, particularly for stress and affect detection. More related work on stress and affect detection, BNNs and active learning can be found in Appendix \ref{section:related_work_appendix}.

\textbf{Scientific contributions:}
\begin{enumerate*}[label=\textbf{(\arabic*)}]
    \item A study of Bayesian deep neural networks for real-time stress and affect detection.
    \item Leveraging the benefits of \textit{Bayesian Active Learning} to model uncertainties, and exploiting several acquisition functions on-device to instantaneously acquire ground truths (affective states) on-the-fly, thereby substantially reducing the labeling load on oracle.
\end{enumerate*}

\section{Our Approach}
\label{section:our_approach}

In this section, we discuss in detail our proposed system and approach to perform Active Learning with physiological data.

\begin{figure}[ht]
\centering
	\includegraphics[width=0.8\linewidth, height=140pt]{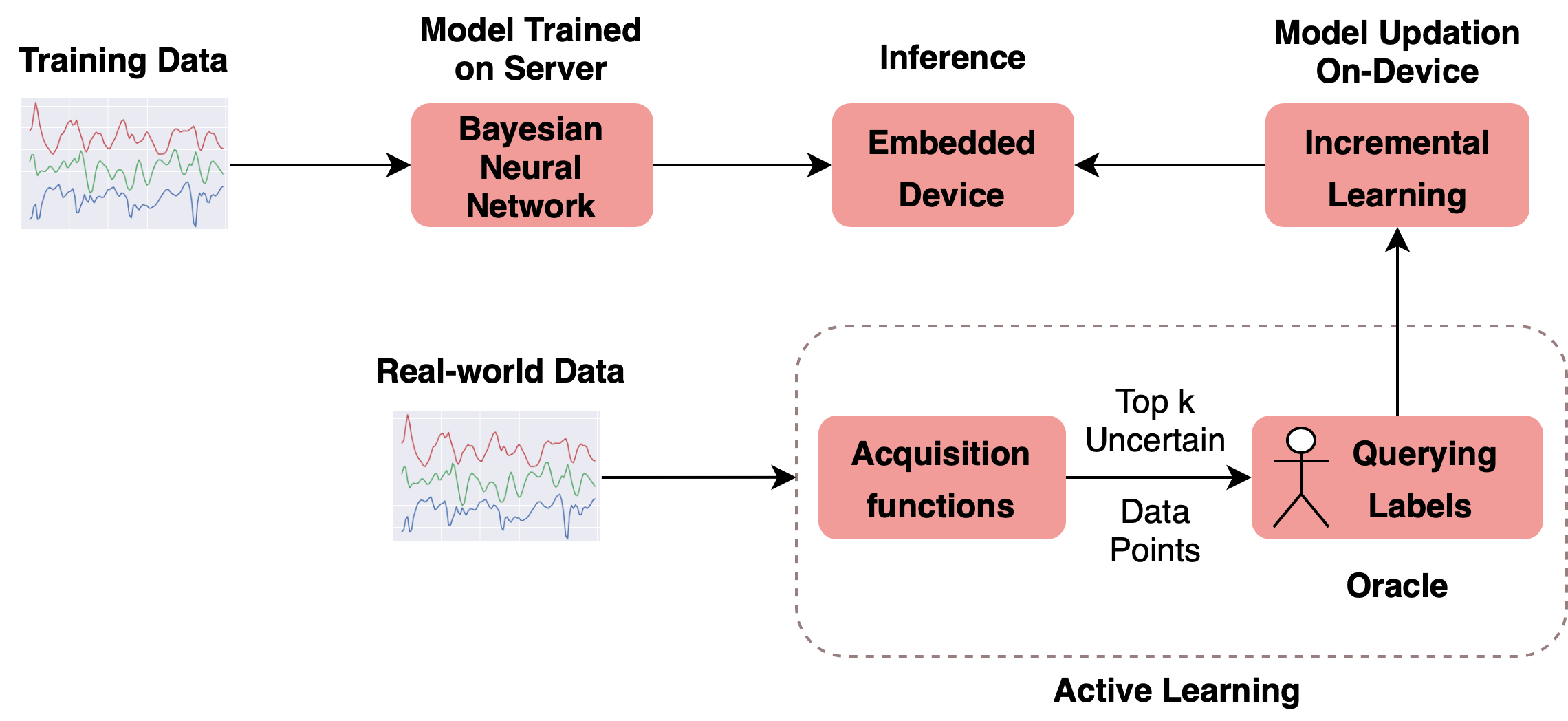}
\caption{Proposed Architecture}
\label{fig:Block_Diagram}
\end{figure}

\subsection{Background on Modeling Uncertainties}
\label{subsection:background}

Bayesian Neural Networks (BNNs) offer a probabilistic interpretation to deep learning models by incorporating Gaussian prior (probability distributions) -- $p(\omega)$ over our model parameters (weights -- $\omega$), thereby modeling output uncertainties. The likelihood model for a classification setting with c classes and x input points is given by,
$$
p(y=c | x, \omega) = softmax(f^\omega(x))
$$
where $f^\omega(x)$ is the model output. However, the posterior distribution of BNNs are not easily tractable, hence it becomes computationally intensive for training and inference.

Gal et al. propose that, \textit{Dropout} -- a stochastic regularization technique \cite{dropout}, can also perform approximate inference over a deep Gaussian process \cite{dropout_yaringal}, and thereby learn to model posterior uncertainties without high computational complexities. This is equivalent to performing Variational Inference (VI), where the posterior distribution is approximated by finding another distribution $q_\theta^*(\omega)$, parameterized by $\theta$, within a family of simplified tractable distributions, while minimizing the Kullback-Leibler (KL) divergence between $q_\theta^*(\omega)$ and the true model posterior $p(\omega | D_{train})$. 

During inference, we can estimate the mean and variance of the BNN's output by applying dropout before every fully-connected layer during train and test time for multiple stochastic passes (\textit{T}). This is equivalent to obtaining predictions and uncertainty estimates respectively from the approximate posterior output of the neural network, thereby making the Bayesian NN \textit{non-deterministic} \cite{dropout_yaringal}. The predictive distribution for a new data point input $x^*$ can be obtained by,
$$
p(y^* | x^*, D_{train}) = \int p(y^* | x^*, \omega) p(\omega | D_{train}) d \omega
$$
where $p(\omega | D_{train}) = q_\theta^*(\omega)$, and $q_\theta^*(\omega)$ is the dropout distribution approximated using VI. Dropout, being a light-weight operation in most existing NN architectures, enables easier and faster approximation of posterior uncertainties.

\subsection{Model Architecture}
\label{subsection:architecture}

In our model, 1D-Convolutional (Conv1D) layers are used for effective temporal extraction of features from physiological data. We use a four-layer Conv1D network with 4, 8, 16 and 32 filters, kernel size 3, and a max-pooling layer of size 2 between each Conv1D layer. This is followed by two Fully-Connected (FC) layers with 32 and 16 neurons each and ReLU activations. An MC-dropout layer with a probability of 0.3 is applied, followed by a softmax layer to obtain the probability scores. The categorical-cross entropy loss of the model is minimized using Adam optimizer.

In order make our model a Bayesian NN so as to obtain uncertainty estimates, we introduce a standard Gaussian prior on the set of our model parameters. Also, to perform approximation inference in our model, we perform dropout at train and test-time as discussed in Section \ref{subsection:background} to sample from the approximate posterior using multiple stochastic forward passes \cite{dropout_yaringal}. After experimenting with multiple dropout iterations (forward passes -- \textit{T}), an optimal \textit{T}=10 is utilized in this paper to determine uncertainties. Effectively, our uncertainty-aware model now is a \textit{Bayesian ensembled Convolutional Neural Network (B-CNN)} which can model uncertainties and be used with existing acquisition functions for AL.

\subsection{Acquisition functions for Active Learning}
\label{subsection:acquisition_functions}

As stated in \cite{BayesianAL}, given a classification model $M$, pool data $D_{pool}$ obtained from real-world, and inputs $x \in D_{pool}$, an acquisition function $a(x, M)$ is a function of $x$ that the active learning system uses to infer the next query point:
$$
    x^* = argmax_{x \in D_{pool}} a(x, M).
$$

Acquisition functions are used in active learning scenarios for approximations in Bayesian CNNs, thereby arriving at the most efficient set of data points to query from $D_{pool}$. We examine the following acquisition functions to determine the most suitable function for on-device computation:

\textbf{Max Entropy:}
Pool points are chosen that maximize the predictive entropy \cite{Max_entropy}.
\begin{align*}
    \mathbb{H}& [y | x, D_{train}] := - \sum_c p(y=c | x, D_{train}) \log p(y=c | x, D_{train})
\end{align*}

\textbf{Bayesian Active Learning by Disagreement (BALD):}
In BALD, pool points are chosen that maximize the mutual information between predictions and model posterior \cite{BALD}. The points that maximize the acquisition function are the points that the model finds uncertain on average, and information about model parameters are maximized under the posterior that disagree the most about the outcome.
$$
\mathbb I[y, \omega | x, D_{train}] =
\mathbb H[y | x, D_{train}] - E_{p(\omega | D_{train})} \big[\mathbb H[y | x, \omega] \big]
$$
where $\mathbb H[y | x, \omega] $ is the entropy of $y$, given model weights $\omega$.

\textbf{Variation Ratios ($VR$):}
The LC (Least Confident) method for uncertainty based pool sampling is performed in $VR$ \cite{Var_Ratios}.
$$
variation-ratio[x] := 1 - \max_y p(y | x, D_{train})
$$

\textbf{Random Sampling:}
This acquisition function is equivalent to selecting a point from a pool of data points uniformly at random.

\section{Experiments and Results}
\label{section:exp_and_results}

\textbf{\textit{SWELL-KW Dataset} \cite{koldijk2014swell}:} The dataset consists of stress data from 25 participants recorded when they were performing knowledge work tasks such as writing reports, making presentations, reading emails, and searching for information. Stress was induced in the participants by stressors such as time-pressure and email interruptions, and their physiological modalities (Heart Rate and Skin Conductance) were recorded. Based on the participants’ self-reports, 3 mental conditions were analyzed -- Neutral (N), Interruption (I) and Time-Pressure (T). We report results for the N vs I\&T classification task.

The train samples are split in random into $D_{train}$ and $D_{pool}$ points, (30-70 ratio) as an approximation of real-world incoming data, while the unseen $D_{test}$ is used for evaluation purposes only in our experiment. To evaluate our Bayesian CNN framework, we initially pre-train our model for a maximum of 10 epochs on a server to establish the baseline efficiencies on the SWELL dataset, and stock the model in the embedded system. To approximate our posterior and obtain predictive uncertainties, we test our BNN model over \textit{T}=10 stochastic iterations, and average our predictions to calculate our final efficiencies. The average initial baseline accuracy without any active learning incorporated is observed to be 79.12\%.

\subsection{Bayesian Active Learning}
\label{section:bayesian_active_learning}

In order to handle ground truth labeling and incorporate model weight updation on incoming test user data, we experiment our proposed framework by deploying the system on a Raspberry Pi 2. The number of acquisition windows used for active learning from $D_{pool}$ can be governed by the \textbf{acquisition adaptation factor} $\eta \in$ [0, 1]. The incremental model updation is simulated for a maximum of 10 epochs for every acquisition iteration, owing to its non-convergence in loss thereafter.

We analyze various AL acquisition functions mentioned in section \ref{subsection:acquisition_functions}, and observe that Variation Ratios ($VR$) acquisition function performs the best, while Random Sampling has the least classification accuracy as expected. In $VR$, only 60\% ($\eta$=0.6) of the  acquisition windows from $D_{pool}$ are required to achieve a test accuracy of 90.38\% from a baseline accuracy of 79.12\%. This is very close to the maximum test accuracy of 91.92\% achieved with $\eta$=1.0 (all $D_{pool}$ windows). Note that, $\eta$ = 0.0 gives the efficiency of the pre-trained model without any data points acquired during incremental learning.

\begin{table}[h!]
\centering
\caption{\textit{Bayesian-CNN} with \textit{SWELL} -- Acquisition Windows ($\eta$) vs Accuracy (\%).}
\label{table:swell_results}
    \begin{tabular}{lllllll}
    \toprule
    \textbf{$\eta$} & \textbf{Max Entropy} & \textbf{BALD} & \textbf{Variation Ratios} & \textbf{Random Sampling}      \\ 
    \midrule
    \textbf{0.0}                    & 79.12           & 79.12           & 79.12            & 79.12   \\
    \textbf{0.2}                    & 82.66           & 81.21           & 83.11            & 81.91   \\
    \textbf{0.4}                    & 86.43           & 86.58           & 88.29           & 86.76   \\
    \textbf{0.6}                    & 89.40           & 90.22           & 90.38           & 88.19   \\
    \textbf{0.8}                    & 89.98           & 90.63           & 90.82           & 88.95   \\
    \textbf{1.0}                    & 91.92           & 91.92           & 91.92           & 91.92   \\
    \bottomrule
    \end{tabular}
\vskip -0.1in
\end{table}

\subsection{On-Device Performance}

Raspberry Pi 2 is used for evaluating our proposed active learning framework as it has similar hardware and software specifications to predominant contemporary IoT devices. We observe an average time of about 0.5 sec is utilized for each stochastic forward pass (\textit{T}) with dropout for acquisition of top $\eta$ windows. Since we perform \textit{T}=10 dropout iterations in our experiment, we observe that an average of about 5 seconds are needed for querying the most uncertain data points. 

\begin{table}[ht]
\centering
\caption{Computation Time Taken for Execution per window}
\label{table:timeOnPi}
\begin{tabular}{lll}
\toprule
\textbf{Process}                & \textbf{Time} \\ 
\midrule
Inference time                  & 9 ms          \\
Time taken per epoch            & 0.6 sec        \\
\bottomrule
\end{tabular}
\end{table}

The model size for SWELL is approximately 115 kB, which is substantially small compared to conventional deep learning models. For real-time deployment feasibility, it is practical to have a threshold (upper limit) on the number of $D_{pool}$ collected in a single acquisition iteration. This can be quantified by either number of windows ($w_a$) or time taken (in seconds).

\section{Conclusion}

This paper presents empirical contributions with emphasis on Bayesian Active Learning for efficient stress and affect detection. First, we benchmark our efficiencies on the SWELL dataset using a Bayesian-CNN model with stochastic dropout, which incorporates information on predictive uncertainties. Second, we perform active learning on-device using the uncertainty approximations by systematically analyzing various acquisition functions for extracting the most informative data points to be queried by the oracle. This can further be extended to other behavior monitoring tasks in mobile and pervasive healthcare. We also believe this work would trigger more research in under explored areas like handling unlabeled data, and active learning for mobile health scenarios.

\bibliography{References}
\bibliographystyle{plain}
\pagebreak

\appendix
\section{Appendix: Related Work}
\label{section:related_work_appendix}

Technological advancements in pervasive and ubiquitous healthcare have drastically improved the quality of human life, and hence has been an actively explored research area. Detecting stress in real-time is crucial for continuous monitoring of mental heath. Deep learning using sensor data has been an evolving domain for computational behavior analysis and healthcare research. Efficient deep learning for stress and affect detection has become the need of the hour, and such approaches have been widely used with promising empirical results by effectively capturing the most discriminative features \cite{stress2}, \cite{stress_wearable}. However, these works do not address the important problem of concept drift, which must be handled by techniques like Active Learning (AL) to combat changes in data distribution over time.

Conventional AL literature \cite{Settles} mostly handle low-dimensional data for uncertainty estimations, but do not generalize to deep neural networks as the data is inherently high-dimensional \cite{BayesianAL}. Deep active learning using uncertainty representations -- which presently are the state-of-the-art AL techniques for high-dimensional data, have had very sparse literature. With the advent of Bayesian approaches to deep learning, Gal et al. \cite{BayesianAL} proposed Bayesian Active Learning for image classification tasks, and is proven to learn from small amounts of unseen data, while Shen et al. \cite{NER_AL} incorporate similar techniques for NLP sequence tagging. However, these techniques are predominantly not discussed for sensor and time-series data.

Incorporating AL for obtaining ground truth in mobile sensing systems has been addressed in a few previous works. Most of them have been addressed with Human Activity Recognition (HAR) tasks \cite{SouravUnlabeled}, \cite{activeharnet}, \cite{ActiveActivity}. Although these works seem to achieve impressive results, the feasibility of on-device learning in mobile health stress and affect detection scenarios with unlabeled data still have not been explored.

\end{document}